\documentclass{article}

\PassOptionsToPackage{numbers, compress}{natbib}

\usepackage[preprint]{neurips_2026}

\usepackage[utf8]{inputenc}
\usepackage[T1]{fontenc}
\usepackage{booktabs}
\usepackage{amsmath}
\usepackage{amssymb}
\usepackage{graphicx}
\usepackage{hyperref}
\usepackage{url}
\usepackage{microtype}
\usepackage{xcolor}

\title{Measuring Maximum Activations in Open Large Language Models}

\author{
    \textbf{Luxuan Chen$^{1,2*}$, Han Tian$^{2,4*}$, Xinran Chen$^{2}$, Rui Kong$^{2}$, Fang Wang$^{2}$}, Jiamin Chen$^{2}$,\\
    \textbf{Yuchen Li$^{2\dagger}$, Jiashu Zhao$^{2}$, Shuaiqiang Wang$^{2}$, Haoyi Xiong$^{3}$, Linghe Kong$^{1}$, Dawei Yin$^{2\dagger}$} \\
    $^1$Shanghai Jiao Tong University \\
    \quad $^2$Baidu Inc. $^3$Independent Researcher $^4$Nankai University \\
    \texttt{yuchenli1230@gmail.com,yindawei@acm.org}
}
\date{}

\begin{document}

\maketitle
\let\thefootnote\relax\footnotetext{$^*$Co-first authors with equal contributions.}
\let\thefootnote\relax\footnotetext{$^\dagger$Corresponding author}
\begin{abstract}
The dynamic range of activations is a first-order constraint for low-bit quantization, activation scaling, and stable LLM inference. Prior work characterized outlier features and massive activations on pre-2024 LLaMA-style models, and the downstream activation-quantization stack inherits that picture without revisiting it for the post-LLaMA open-model boom. We ask the deployment-oriented question: how \emph{large} can activations get in modern open LLMs, and how does this magnitude vary across families, generations, and training stages? Under a unified pipeline (5{,}000-sample multi-domain corpus, family-specific tokenization, identical hooks across embeddings, hidden states, attention, MLP/MoE, SwiGLU gates, and final norm), we measure global and layerwise maxima on 27 checkpoints from 8 open families spanning dense, MoE, vision-language, intermediate-training, and instruction-tuned variants. We find that (i) global maxima span over nearly four orders of magnitude at comparable parameter counts, with Qwen3.5 and MoE checkpoints in the $10^2$--$10^3$ range and Gemma3-27B-it reaching $\sim\!7\!\times\!10^5$; (ii) cross-family and cross-generation comparisons break simple monotonic scaling; and (iii) MoE checkpoints exhibit $14.0$--$23.4\times$ lower peaks than matched-scale dense counterparts, while the residual stream carries the global maximum in 22/24 checkpoints. A lightweight INT-8 sanity check shows that measured maxima co-vary with low-bit reconstruction error via activation-scale selection. We conclude that maximum activation magnitude is a model property tied to family, architecture, and training stage---not a simple byproduct of size---and should be measured and reported alongside any open-weight release before low-bit deployment. The code is publicly available at \url{https://github.com/clx1415926/Max_act_llm}.
\end{abstract}

\section{Introduction}
\label{sec:introduction}

The activation dynamic range of a large language model (LLM) is not merely a descriptive statistic: it determines the numerical range that inference systems, activation quantizers, and scaling rules must accommodate \citep{deepseekai2025deepseekv3technicalreport}. In low-bit inference, for example, a per-tensor activation scale is often chosen to cover the largest magnitude observed on a calibration set. A small number of extremely large activations can therefore dominate the scale, waste most quantization levels on rarely used values, and amplify reconstruction error for ordinary activations. This paper studies a simple but deployment-critical quantity: the \emph{maximum activation magnitude}, defined as the largest absolute activation observed across layers and key components under a fixed evaluation protocol.

Extreme activations, massive activations, and outlier features have been studied from several perspectives, including existence tests, token- or feature-level localization, and functional interventions. However, the deployment question remains less systematically mapped: how large can activations become in recent open LLMs, where do the largest values appear, and how do they change with model family, architecture, model generation, and training stage? This question is increasingly important because modern open models no longer differ only in parameter count. They vary in normalization and training recipes, dense versus MoE computation, vision-language adaptation, instruction tuning, and released intermediate training stages. As a result, parameter scale alone may be an unreliable proxy for activation range.

Prior work on extreme activations falls along two largely separate lineages. The \emph{interpretability} line begins with \citep{dettmers2022llmint8}, who defined \emph{emergent outlier features} via a $6\sigma$ rule on OPT/BLOOM, and \citep{sun2024massive}, who introduced \emph{massive activations} as coordinates that are simultaneously large ($|x_i|>100$) and locally sparse ($\geq 1000\times$ the per-token median); \citep{bondarenko2023quantizable} attributed them to attention heads needing a ``no-op'' route through the residual stream, and very recent work refines this picture---\citep{gu2025attentionsink} trace when attention sinks emerge during pretraining, and \citep{zhu2026spike} decouple massive-activation ``spikes'' from sinks and localize them to early-layer step-up blocks under pre-norm transformers. All of these studies treat the phenomenon categorically and analyze a handful of LLaMA-family or single-architecture checkpoints. The \emph{quantization} line treats the same activations as a deployment obstacle: SmoothQuant~\citep{xiao2023smoothquant}, AWQ~\citep{lin2024awq}, GPTQ~\citep{frantar2023gptq}, and Outlier Suppression+~\citep{wei2023outliersuppression} migrate or rescale outlier mass; rotation methods QuaRot~\citep{ashkboos2024quarot}, SpinQuant~\citep{liu2024spinquant}, and DuQuant~\citep{lin2024duquant} remove the outlier basis; FlatQuant~\citep{sun2025flatquant} learns affine flattening transforms; PrefixQuant~\citep{chen2024prefixquant} and KIVI~\citep{liu2024kivi} target the KV cache; and FP8 pretraining pipelines~\citep{deepseekai2025deepseekv3technicalreport,deepseekv3hardware} fold analogous mitigations into low-precision training, with~\citep{lingscaling} arguing that high-sparsity MoE routing further changes the activation regime. All of these mitigations \emph{transform away} the upper bound rather than measuring how it varies across modern releases.

It remains unclear whether either discovery still holds for the recent wave of post-LLaMA open releases---Qwen2.5/3/3.5~\citep{qwen25report,qwen3report}, Qwen2.5-VL~\citep{qwen25vl}, Gemma~2 and Gemma~3~\citep{gemma2report,gemma3report}, the Ling-mini series~\citep{lingreport}, and gpt-oss~\citep{gptoss2025}---which diverge from earlier LLaMA-style models along multiple axes simultaneously: normalization stack, gated MLP variants, MoE routing, multimodal adaptation, intermediate-training releases, and instruction tuning. To our knowledge no prior study reports activation magnitudes across these families under a unified protocol. Our paper is complementary along both axes: we provide the first unified-protocol measurement of the global maximum $M=\max|a|$ across $27$ post-LLaMA open checkpoints from $8$ families, treat $M$ as a continuous releasable model property rather than a binary outlier flag, and connect $M$ directly to per-tensor INT-8 reconstruction error---inputs that the mechanistic and quantization-mitigation lines currently lack.

We address this gap with a unified empirical survey of maximum activations in modern open LLMs. Our main analysis covers 24 checkpoints from 8 model families: Qwen2.5, Qwen2.5-VL, Qwen3, Qwen3.5, Gemma2, Gemma3, Ling, and GPT-OSS. We additionally analyze 3 Qwen2.5-Instruct checkpoints to isolate the effect of supervised fine-tuning. All checkpoints are evaluated on the same 5,000-sample multi-domain corpus, with the text corpus re-tokenized for each model family. During forward inference, we use PyTorch hooks to stream activation statistics from embeddings, layerwise hidden states, attention outputs, MLP or MoE outputs, SwiGLU gate pre-activations, and final normalization outputs. This protocol lets us compare global maxima, layerwise peak trajectories, carrier components, family and generation effects, and matched architectural or training contrasts under the same measurement pipeline.

\textbf{Contributions.} Our work makes the following contributions:
\begin{itemize}\itemsep1pt \parskip0pt \topsep2pt
    \item \textbf{Largest-to-date cross-family activation survey.} We measure global and layerwise maximum activations on 24 checkpoints from 8 modern open families (Qwen2.5/2.5-VL/3/3.5, Gemma2/3, Ling, GPT-OSS)---spanning dense, MoE, vision-language, intermediate-training, and instruction-tuned variants---under a single unified pipeline (5,000-sample multi-domain corpus, family-specific tokenization, identical hooks across embeddings, hidden states, attention/MLP/MoE outputs, SwiGLU gates, and final norm), moving beyond the LLaMA-derivative monoculture of~\citep{dettmers2022llmint8,sun2024massive}.
    \item \textbf{Continuous magnitude reformulation of ``massive activation.''} We replace the binary same-token criterion of~\citep{sun2024massive} with the deployment-relevant statistic $M=\max|a|$, and show the two views can disagree---some checkpoints failing the binary criterion are the easiest to quantize, while some passing it are the hardest.
   \item \textbf{Five matched-design comparisons.} We isolate (i) within-family scaling, (ii) same-scale MoE-vs-dense, (iii) same-family vision-language-vs-text-only, (iv) same-backbone Base-vs-Instruct, and (v) same-family training-stage effects on the same measurement substrate, providing the first observational decomposition of scale, family, generation, architecture, modality, and training progress for activation peaks.
    \item \textbf{Empirical findings with deployment implications.} (a) Global maxima vary by orders of magnitude across families at comparable parameter counts and break simple monotonic scaling; (b) the residual stream carries the global maximum in 22/24 main checkpoints; (c) MoE reduces peak magnitudes by $14.0$--$23.4\times$ relative to nearby dense counterparts; (d) SFT mainly compresses late-layer peaks; (e) training progress can monotonically increase the global maximum at fixed family and architecture; and (f) a lightweight per-tensor INT-8 probe shows higher $M$ correlates with substantially lower SQNR.
    \item \textbf{Open pipeline and per-checkpoint statistics.} We release the hook-based measurement code and per-checkpoint activation statistics to support reproducibility and future quantization, scaling, and architecture research. We do \emph{not} claim a causal mechanism for the observed differences (see Section~\ref{sec:discussion}).
\end{itemize}

\section{Measurement Protocol}
\label{sec:experimental-setup}

\begin{figure}[!htbp]
\centering
\includegraphics[width=.99\linewidth]{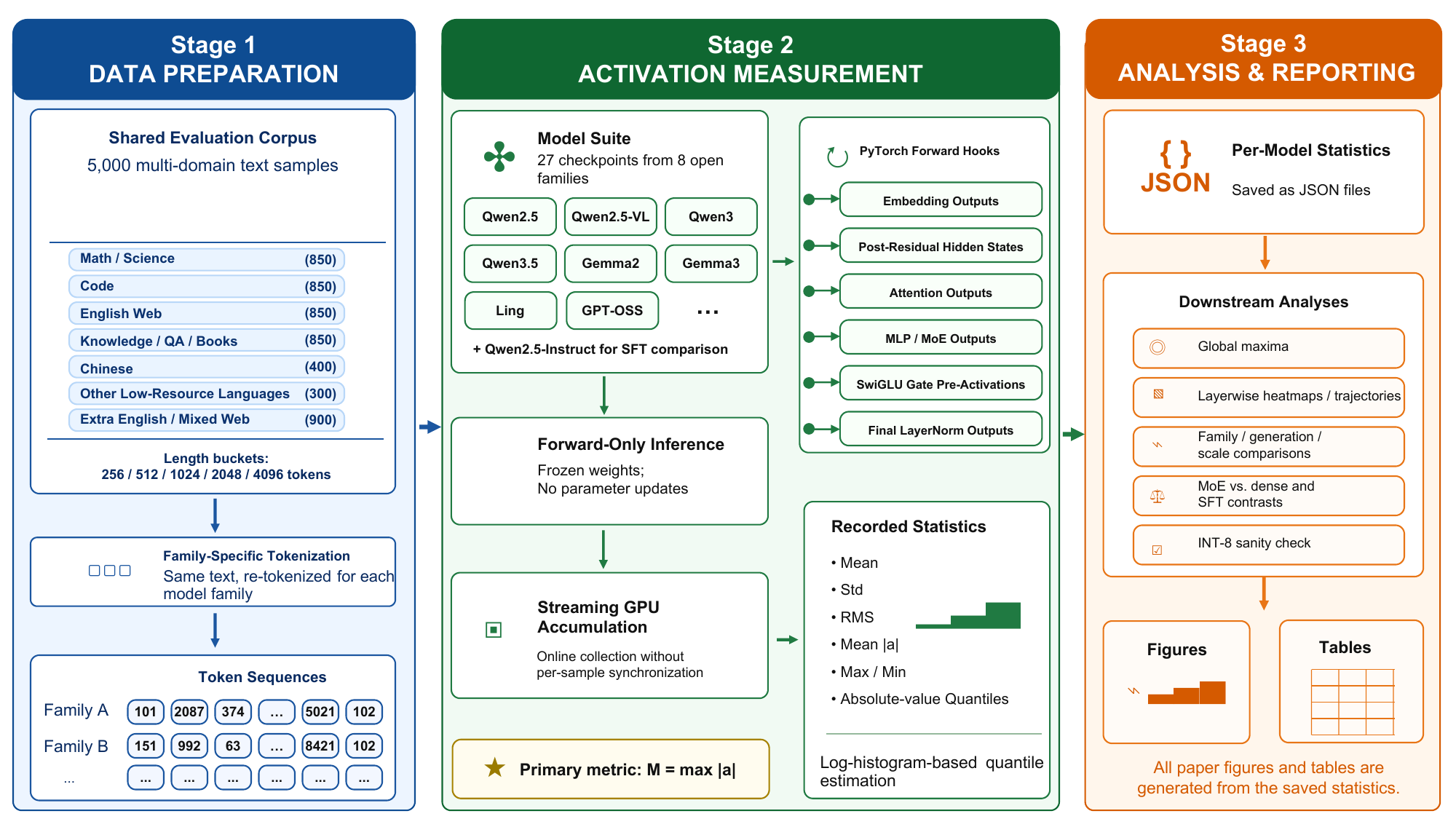}
\caption{Overview of the pipeline.}
\label{fig:overview}
\end{figure}

The overall pipeline is shown in Figure \ref{fig:overview}, which consists of three steps: Data Preparation, Activation Measurement, and Analysis. We use a unified offline evaluation protocol. We first construct a multi-domain text corpus, tokenize the same text with each model family's tokenizer, run forward inference on each checkpoint, and record layerwise activation statistics. All figures and tables are generated from the resulting per-model statistics.

\subsection{Corpus construction}

The target evaluation corpus contains 5,000 samples. The data are sampled from RedPajama \citep{shen2024slimpajamadcunderstandingdatacombinations} sources and bucketed by content type. The target category counts are 850 mathematical or scientific samples, 850 code samples, 850 English web samples, 850 knowledge-oriented samples such as encyclopedic, book, or Q\&A text, 400 Chinese samples, 300 samples in other low-resource languages, and 900 additional English or mixed web samples. This design reduces the risk that maximum-activation statistics are dominated by a single domain and ensures that the corpus covers formal text, natural web text, knowledge-intensive text, code, and multilingual content.

The corpus also controls sequence-length diversity. Samples are randomly truncated to 256, 512, 1024, 2048, or 4096 tokens with target proportions of 1\%, 1\%, 2\%, 3\%, and 93\%, respectively. The corpus is therefore dominated by long-context inputs while retaining a small number of short and medium-length sequences. The resulting corpus has an average length of approximately 3899 tokens, corresponding to roughly 19.5M tokens in total.

To avoid tokenizer mismatch artifacts, the text corpus is held fixed while tokenization is performed separately for each model family. Thus, models receive semantically identical text but token sequences aligned with their own tokenizer, reducing activation-statistics bias caused by tokenizer incompatibility.

\subsection{Model suite and instrumentation}

We select models according to three principles. First, we cover recent mainstream open LLM families rather than restricting the study to earlier LLaMA-style models. Second, we include multiple parameter scales and architectural forms, allowing us to separate the effects of scale, family, and architecture. Third, we include special variants such as MoE models, vision-language models, intermediate training checkpoints, and instruction-tuned models, so that we can examine whether maximum activations change with routing, modality adaptation, training progress, or supervised fine-tuning (SFT).

The main experiment contains 24 checkpoints from 8 families: Qwen2.5, Qwen2.5-VL, Qwen3, Qwen3.5, Gemma2, Gemma3, Ling, and GPT-OSS. Except for the publicly released Gemma3 checkpoints, which are instruction-tuned models, we treat the main-analysis checkpoints as base or intermediate-training checkpoints, as Shown in Table \ref{tab:model-families}. Therefore, the Gemma2/Gemma3 comparison should be interpreted as a public-checkpoint family-level contrast rather than a strict base-to-base generational ablation.

\subsection{Recorded statistics and peak stability}

The statistics pipeline has three stages. First, the shared text corpus is converted into token sequences with each model family's tokenizer. Second, each checkpoint is loaded with its full weights and evaluated with forward inference only; no parameters are modified. During inference, PyTorch forward hooks collect six classes of activation tensors: embedding outputs, layerwise hidden states after residual updates, layerwise attention outputs, layerwise MLP outputs or MoE block outputs, MLP gate pre-activations in SwiGLU-style architectures, and final LayerNorm outputs. Third, per-model JSON statistics are used to generate all figures and tables. For each captured component, we record the mean, standard deviation, RMS, mean absolute value, maximum value, minimum value, and streaming estimates of absolute-value quantiles.

Because the global maximum activation is an extreme statistic, we first verify that it is not triggered by a small number of accidental samples. For four representative models, we construct category-proportional subsamples of 1,000 and 2,000 examples from the original 5,000-example corpus. Each subsample size is repeated 5 times, and each repeat runs the full activation scan. The resulting peaks consistently reproduce the order of magnitude of the 5k reference run. The largest coefficient of variation across 1k repeats is 10.1\% for Qwen3-30B-A3B, and the largest coefficient of variation across 2k repeats is 8.2\%. These results indicate that the reported maximum activations are not accidental single-sample artifacts and that the measurements are statistically robust at the scale studied here.

\section{From Binary Massive Activations to Continuous Peaks}
\label{sec:existence}

The empirical story proceeds from definition to mechanism to comparison. We first connect our deployment-oriented maximum $M$ to the binary massive-activation criterion used in prior work, then ask where the largest values are carried, and finally compare families, generations, architectures, and training stages under matched designs.

\subsection{Relationship to the Sun criterion}

Our main metric is the global maximum activation, $M=\max |a|$, taken across all six hooked component classes (embeddings, layerwise hidden states, attention outputs, MLP/MoE outputs, SwiGLU gate pre-activations, final LayerNorm) and all layers; this is the value plotted in every bar chart and used in every matched-pair ratio in Sections~\ref{sec:family-scale-generation} and~\ref{sec:special-architectures}. The Top-$k$ values reported in Table~\ref{tab:massive-existence} are drawn from a single representative layer chosen for the local-sparsity diagnostic and may therefore differ from $M$ when a different layer carries the global peak. Before turning to this macro-level magnitude analysis, we first drill down into whether the global extrema also satisfy a commonly used local sparsity definition from prior work. We adopt the same-token criterion of \citep{sun2024massive}: given a hidden state vector $x\in\mathbb{R}^{d}$ for one token, a coordinate $x_i$ is counted as a massive activation if it simultaneously satisfies $|x_i| > 100$ and $\frac{|x_i|}{\operatorname{median}_{j=1}^{d}|x_j|} > 1000$. At the model level, a checkpoint passes the criterion if any hidden layer contains at least one token-feature coordinate satisfying both thresholds.

\subsection{Overall existence and failure mechanisms}

Table~\ref{tab:massive-existence} summarizes representative activation locations for each checkpoint based on the full layerwise scan. Overall, 20 of the 24 main-analysis checkpoints pass the Sun criterion, indicating that massive activations remain widespread in recent open LLMs.

\begin{table}[t]
\centering
\scriptsize
\setlength{\tabcolsep}{3pt}
\caption{Representative activation locations for the 24 main-analysis checkpoints. We report the five largest absolute activations, Top-10, Top-100, Top 1\%, Top 10\%, and median values. The representative location is selected from the full layerwise scan: for passing models, we use a hidden layer containing a same-token sparse extreme value; for failing models, we use the full-model peak location. The median is the median absolute value across the $d$ dimensions of the single token activation vector carrying the reported peak. Models marked with $\times$ fail the Sun criterion due to insufficient local ratio or insufficient absolute magnitude.}

\resizebox{\linewidth}{!}{%
\begin{tabular}{lrrrrrrrrrr}
\toprule
Model & Top 1 & Top 2 & Top 3 & Top 4 & Top 5 & Top-10 & Top-100 & Top 1\% & Top 10\% & median \\
\midrule
Qwen2.5-1.5B$^{\times}$        & 7,968 & 7,968 & 7,968 & 7,968 & 7,968 & 7,968 & 7,776 & 3.72 & 1.37 & 13.9 \\
Qwen2.5-7B         & 13,248 & 13,248 & 13,248 & 13,248 & 13,248 & 13,184 & 12,224 & 3.94 & 1.37 & 5.00 \\
Qwen2.5-32B        & 22,144 & 21,888 & 21,760 & 21,760 & 21,760 & 21,632 & 21,120 & 12.9 & 3.94 & 8.62 \\
\midrule
Qwen2.5-VL-3B      & 3,248 & 3,248 & 3,248 & 3,248 & 3,088 & 3,056 & 2,944 & 6.05 & 1.97 & 1.84 \\
Qwen2.5-VL-7B      & 8,256 & 8,256 & 8,256 & 8,256 & 8,256 & 8,256 & 7,552 & 5.57 & 1.91 & 3.23 \\
Qwen2.5-VL-32B     & 22,144 & 22,144 & 22,144 & 22,144 & 22,144 & 22,144 & 21,248 & 14.6 & 4.37 & 9.31 \\
\midrule
Qwen3-1.7B         & 14,208 & 14,144 & 14,144 & 14,144 & 14,144 & 14,080 & 13,952 & 103.9 & 52.5 & 4.19 \\
Qwen3-8B           & 17,664 & 17,664 & 17,664 & 17,664 & 17,664 & 17,664 & 16,896 & 4.05 & 1.32 & 3.14 \\
Qwen3-30B-A3B      & 1,512 & 1,512 & 1,512 & 1,512 & 1,512 & 1,512 & 1,392 & 0.348 & 0.099 & 0.578 \\
Qwen3-32B          & 35,328 & 35,328 & 35,328 & 35,328 & 35,328 & 35,328 & 34,304 & 11.7 & 3.61 & 12.4 \\
\midrule
Qwen3.5-0.8B$^{\times}$        & 122.0 & 122.0 & 114.5 & 114.5 & 114.0 & 22.5 & 3.23 & 12.2 & 6.28 & 2.20 \\
Qwen3.5-9B$^{\times}$          & 956.0 & 920.0 & 920.0 & 912.0 & 908.0 & 848.0 & 87.5 & 12.1 & 5.48 & 3.42 \\
Qwen3.5-27B        & 167.0 & 167.0 & 167.0 & 167.0 & 160.0 & 152.0 & 127.0 & 0.965 & 0.445 & 0.225 \\
Qwen3.5-35B-A3B$^{\times}$     & 132.0 & 131.0 & 129.0 & 126.5 & 125.0 & 121.0 & 40.8 & 1.37 & 0.595 & 0.719 \\
\midrule
gemma-2-2b           & 2,992 & 2,992 & 2,976 & 2,944 & 2,944 & 2,848 & 884.0 & 6.71 & 2.34 & 50.5 \\
gemma-2-9b         & 1,656 & 1,560 & 1,160 & 1,112 & 1,040 & 808.0 & 294.0 & 16.4 & 6.11 & 1.50 \\
gemma-2-27b        & 157,696 & 156,672 & 155,648 & 153,600 & 151,552 & 149,504 & 83,456 & 564.7 & 298.5 & 122.0 \\
\midrule
gemma-3-4b-it      & 245,760 & 243,712 & 242,688 & 241,664 & 241,664 & 237,568 & 165,888 & 828.7 & 461.2 & 86.5 \\
gemma-3-27b-it     & 696,320 & 696,320 & 696,320 & 696,320 & 696,320 & 696,320 & 667,648 & 602.4 & 316.6 & 31.2 \\
\midrule
Ling-mini-5T       & 7,648 & 7,648 & 7,648 & 7,648 & 7,648 & 7,648 & 1,976 & 16.8 & 7.03 & 3.59 \\
Ling-mini-10T      & 9,024 & 9,024 & 9,024 & 9,024 & 9,024 & 9,024 & 3,296 & 9.59 & 3.14 & 3.39 \\
Ling-mini-15T      & 9,600 & 9,600 & 9,600 & 9,600 & 9,600 & 9,600 & 8,192 & 10.4 & 3.31 & 4.06 \\
Ling-mini-20T      & 10,240 & 10,048 & 9,984 & 9,920 & 9,856 & 9,664 & 8,512 & 3.52 & 1.05 & 1.32 \\
\midrule
gpt-oss-20b        & 43,008 & 42,496 & 42,496 & 42,496 & 42,496 & 42,496 & 42,240 & 48.3 & 12.0 & 5.94 \\
\bottomrule
\end{tabular}%
}

\label{tab:massive-existence}
\end{table}

\begin{figure}[!htbp]
\centering
\includegraphics[width=.9\linewidth]{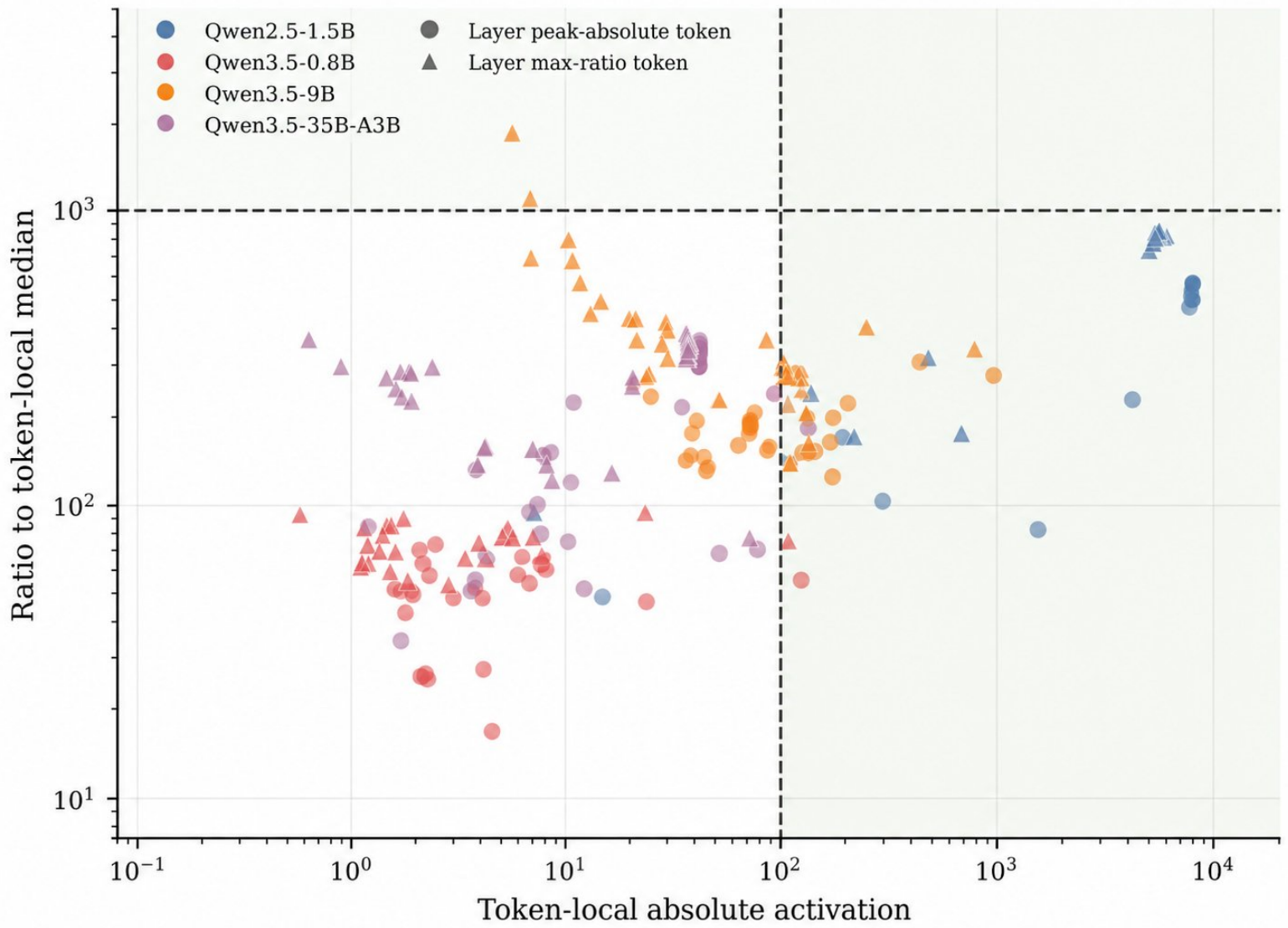}
\caption{Failure modes for the four checkpoints that do not satisfy the Sun criterion. Colors indicate models; circles denote each layer's peak token, and triangles denote each layer's highest local-ratio token. Dashed lines mark the absolute-magnitude and local-ratio thresholds; the upper-right region is the passing region. No point overlaps this region.}
\label{fig:failure-scatter}
\end{figure}

The four failing checkpoints reveal two distinct failure mechanisms. Qwen2.5-1.5B reaches an absolute peak of 7,968, but the median absolute value within the peak token is 13.9, giving a local ratio of roughly 574 and falling below the $1000\times$ threshold. This model therefore exhibits large but relatively dense activations rather than locally sparse massive activations. In contrast, Qwen3.5-0.8B, Qwen3.5-9B, and Qwen3.5-35B-A3B fail because their overall activation scale is systematically suppressed. Figure~\ref{fig:failure-scatter} shows that all failing points avoid the upper-right passing region. This diagnosis confirms that local ratio alone does not fully characterize activation-range risk, motivating our use of the global absolute maximum as the primary metric for quantization and deployment analysis.

We retain the absolute peak, the local-ratio scatter (Figure~\ref{fig:failure-scatter}), and the dual reporting in Table~\ref{tab:massive-existence} as the diagnostic for these failures rather than introducing a normalization-stack-specific re-analysis. The diagnostic is consistent with the architectural account of \citep{zhu2026spike}, in which residual-stream spikes are generated by a small number of early-layer step-up blocks and shaped by the pre-norm normalization stack; for the deployment-oriented question of this paper---``how large can the activation magnitude become''---the absolute peak $M$ is the directly relevant quantity. We therefore treat the binary criterion mainly as a descriptive bridge to prior work and report $M$ as the primary metric throughout the paper.

\section{Where Maximum Activations Form}
\label{sec:layerwise}

\subsection{Layerwise intensity distribution}

\begin{figure}[!htbp]
\centering
\includegraphics[width=.92\linewidth]{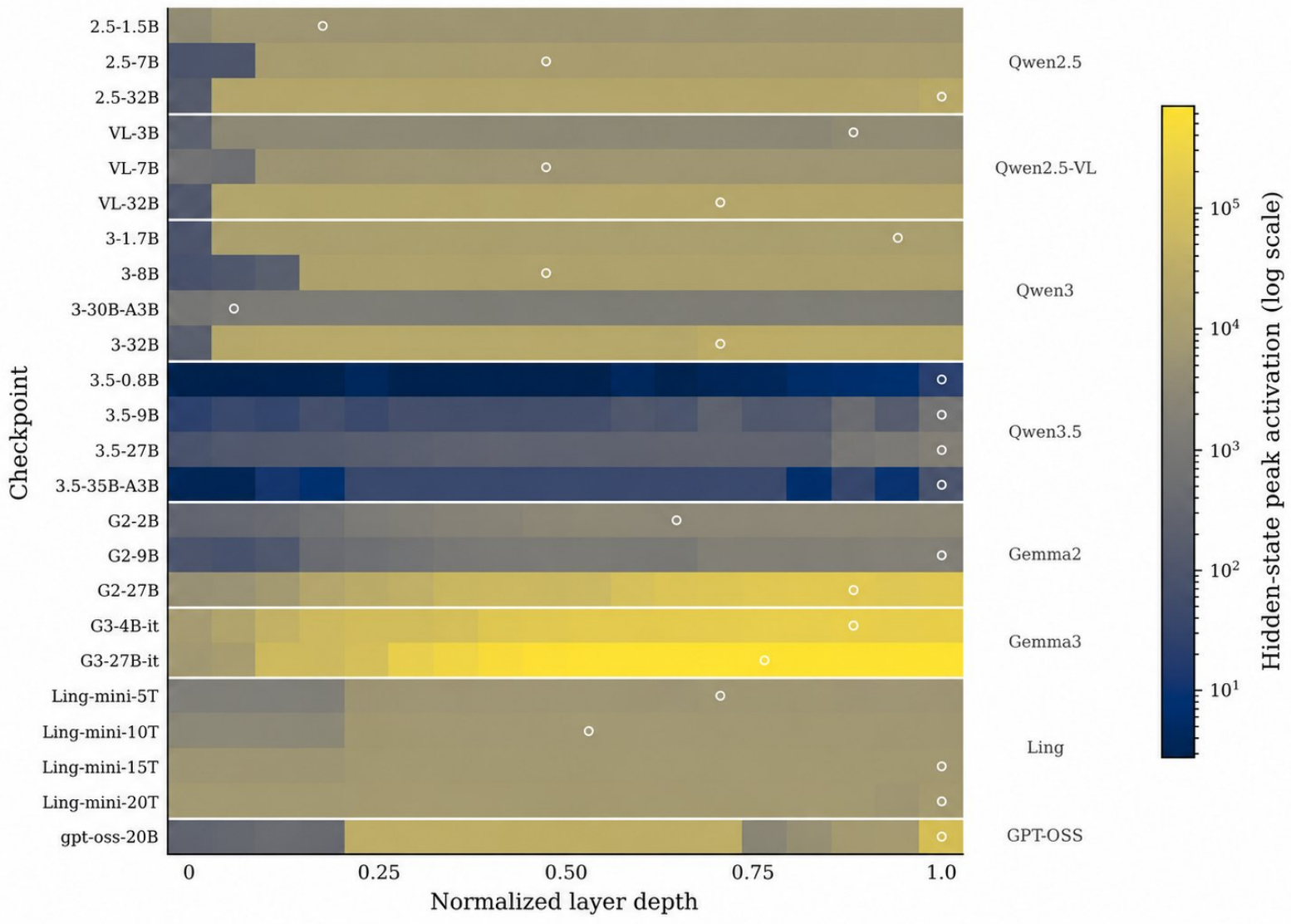}
\caption{Layerwise heatmap of hidden-state peak magnitudes. The horizontal axis is normalized depth, and color indicates layerwise absolute peak magnitude on a log scale. White hollow markers indicate the peak depth bin for each checkpoint.}
\label{fig:layer-heatmap}
\end{figure}

Figure~\ref{fig:layer-heatmap} shows the normalized-depth heatmap of hidden-state peak magnitudes for all main-analysis checkpoints. Peak depth has no universal location across architectures: even within the same family, maxima can occur in shallow, middle, or deep layers. Therefore, reporting only the peak layer index is less informative than characterizing the full layerwise trajectory, namely how peak magnitudes accumulate, jump, plateau, or decay with network depth.

\subsection{Two layerwise patterns}

\begin{figure}[!htbp]
\centering
\includegraphics[width=.98\linewidth]{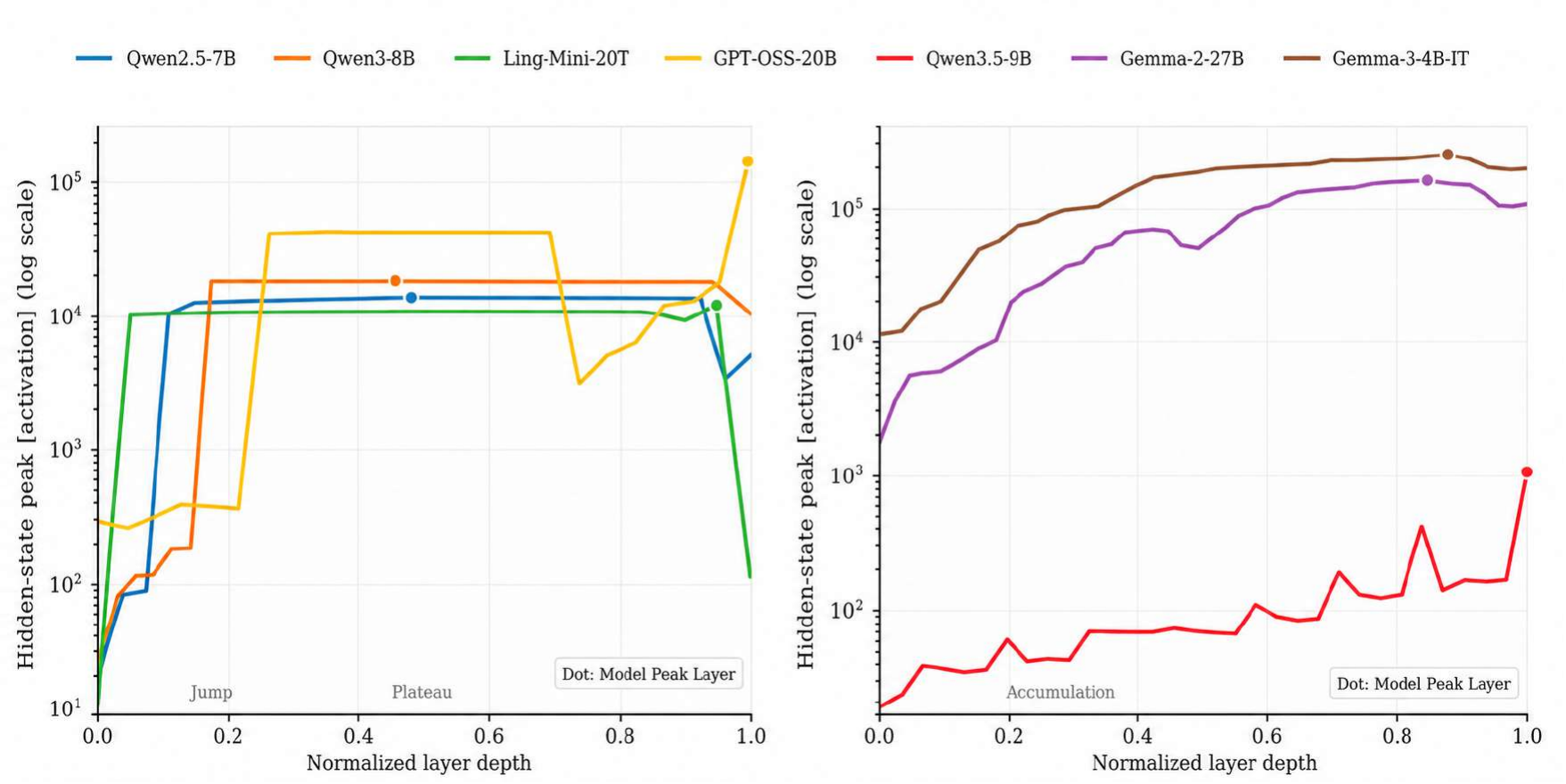}
\caption{Representative layerwise trajectories for the two main emergence patterns. Left: jump-and-plateau models rise sharply in early or middle layers and remain high afterward. Right: gradual-accumulation models grow more smoothly with depth and often peak in later layers. Each curve corresponds to one representative checkpoint; markers indicate the peak layer.}
\label{fig:emergence-main}
\end{figure}

The layerwise trajectories broadly fall into two patterns, illustrated in Figure~\ref{fig:emergence-main}. The first is a jump-and-plateau pattern: activation magnitude rises sharply in early or middle layers and then remains high over a long layer interval, as in Qwen2.5 and GPT-OSS. The second is a gradual-accumulation pattern: activation magnitude increases more smoothly with depth and often reaches its maximum in later layers, as in Qwen3.5 and Gemma. This distinction indicates that maximum activations are governed not only by the physical depth of the peak layer, but also by the dynamics through which the peak forms. The pattern is strongly associated with model family and architecture rather than being a monotonic function of parameter scale.

We treat this dichotomy as a qualitative description of the depth-normalized layerwise heatmap in Figure~\ref{fig:layer-heatmap} and the representative trajectories in Figure~\ref{fig:emergence-main}, not as a quantitative classifier. The two patterns are consistent with the architectural account of \citep{zhu2026spike}, in which spike formation is concentrated in a small number of early-layer step-up blocks under pre-norm transformers, and the cross-family heatmap already shows the relevant separation; introducing a numerical classifier on $n=24$ trajectories would not change the matched-design contrasts in Sections~\ref{sec:family-scale-generation}--\ref{sec:special-architectures}, which use $M$ rather than the trajectory shape.

\subsection{Carrier components}

Across the 24 main-analysis checkpoints, 22 global maxima occur in layerwise hidden states. GPT-OSS-20B is a component-level exception whose global maximum comes from the MLP output, and the failing Qwen3.5-0.8B checkpoint peaks at the final LayerNorm output. If we restrict attention to the 20 checkpoints that satisfy the local sparsity criterion, all qualifying coordinates occur in layerwise hidden states. The residual stream is therefore the dominant carrier through which extreme activation magnitudes are propagated and preserved.

\section{What Controls Peak Magnitude?}
\label{sec:family-scale-generation}

\subsection{Within-family scaling}

\begin{figure}[!htbp]
\centering
\includegraphics[width=.96\linewidth]{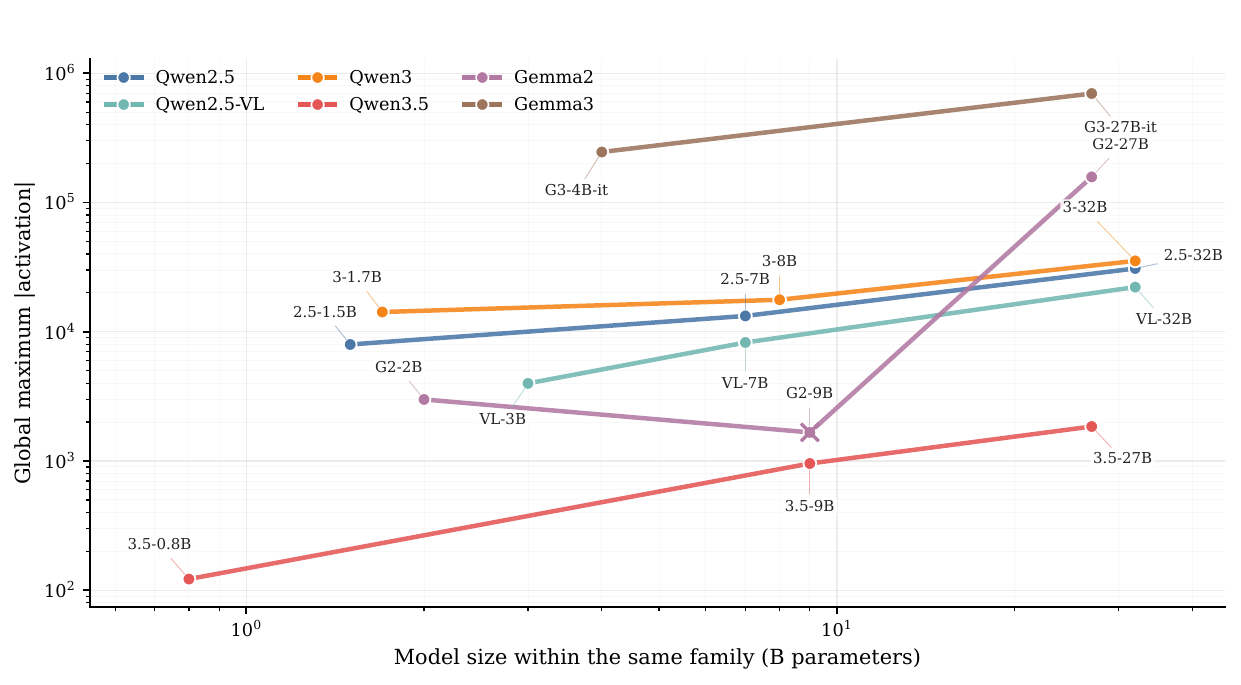}
\caption{Within-family scaling effects. The figure compares size changes only within the same family and model form; crosses mark local non-monotonic cases. The vertical axis is global maximum activation magnitude on a log scale.}
\label{fig:family-scaling-non-moe}
\end{figure}

Figure~\ref{fig:family-scaling-non-moe} compares checkpoints of different sizes within the same family and model form. Most families, including Qwen2.5, Qwen3.5, and Gemma3, show a stable within-family scale effect: the global maximum activation increases with parameter count. Gemma2 is the main local non-monotonic exception, with the 9B checkpoint peaking below the 2B checkpoint before the 27B checkpoint rises again. These results suggest that, when model family and training form are fixed, model size often amplifies activation extremes, although individual checkpoints can still deviate due to training or recipe differences.

\subsection{Cross-family magnitude differences}

\begin{figure}[!htbp]
\centering
\includegraphics[width=.98\linewidth]{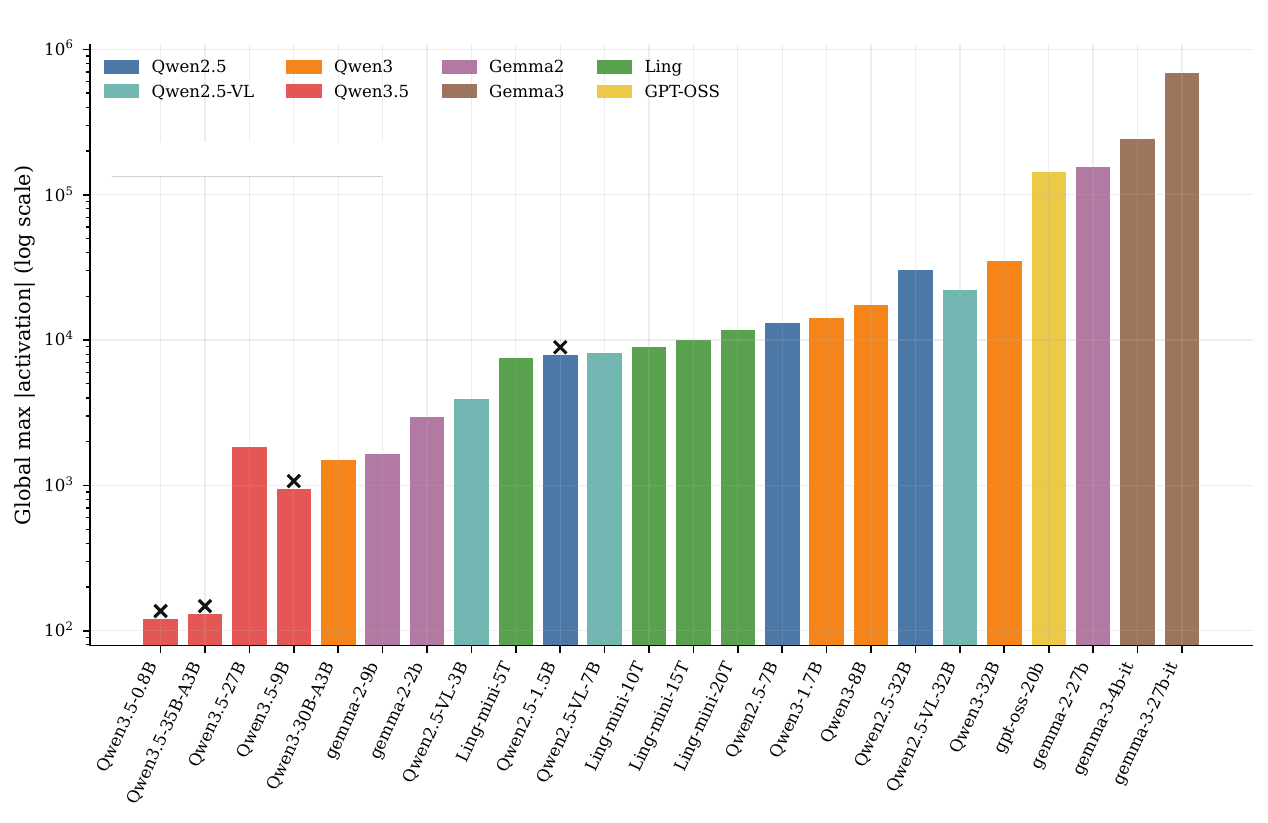}
\caption{Global maximum activation magnitudes for the 24 main-analysis checkpoints. The vertical axis is on a log scale. A $\times$ marker indicates a model that does not satisfy the massive-activation existence criterion.}
\label{fig:peak-landscape}
\end{figure}

Figure~\ref{fig:peak-landscape} summarizes the global maximum activation magnitudes of all 24 main-analysis checkpoints. Cross-family variation is much larger than within-family scaling variation, spanning several orders of magnitude. For example, Qwen3.5 is concentrated in a low-magnitude regime around hundreds to low thousands, whereas Gemma3-27B-it reaches a global maximum of 696,320. This directly shows that the severity of maximum activations is strongly reshaped by family-level architecture and training choices.

\subsection{Non-monotonic generational evolution}

Appendix Figure~\ref{fig:generation-same-scale} compares generational trends at similar model sizes. Maximum activation magnitude does not monotonically shrink or grow with release time; instead, it is highly family-dependent. Across three size groups, Qwen exhibits an inverted-V trajectory: maximum activations increase from Qwen2.5 to Qwen3 and are then strongly suppressed in Qwen3.5. In contrast, Gemma shows a sharp increase from Gemma2 to Gemma3 across both matched size groups.

The detailed same-scale generational plot is moved to Appendix Figure~\ref{fig:generation-same-scale} to keep the main-text submission within the nine-page content budget.

\subsection{Empirical rule of thumb}

Taken together, these measurements suggest a practical rule of thumb. Within a fixed family, increasing parameter count often increases maximum activation magnitude. Once family or generation changes, however, design and training differences can override this monotonicity. For inference and quantization boundaries, family identity and generation are therefore at least as important as parameter count. The appendix extends this logic to matched MoE-vs-dense, vision-language-vs-text, Base-vs-Instruct, and training-stage contrasts.

\section{Deployment Takeaways and Conclusion}
\label{sec:main-conclusion}

The preceding sections establish a compact logic for deployment: define the relevant upper-bound statistic, verify that it is not reducible to a binary outlier label, locate its carrier, and then compare matched model factors. This logic supports three takeaways. First, maximum activation magnitude is a family- and architecture-dependent model property: MoE routing, modality adaptation, instruction tuning, and training stage all shift either the global peak or its layerwise carrier. Second, the residual stream is the dominant carrier of extreme values, so activation quantization and scaling policies should inspect hidden-state peaks rather than only attention or MLP outputs. Third, the INT-8 probe in Appendix~\ref{sec:discussion} shows that larger measured peaks can translate into lower reconstruction SQNR through scale selection, making $M$ a practical model-card statistic rather than only a descriptive outlier measure.

In summary, $M=\max|a|$ is not predicted by parameter count alone: within-family scaling often increases $M$, but cross-family, cross-generation, and cross-architecture comparisons break monotonic trends by orders of magnitude. Detailed matched-factor analyses, quantization checks, limitations, and supplementary figures are provided in the appendix.

\bibliographystyle{plain}
\bibliography{reference,reference-extension}

\newpage
\appendix

\section{Supplementary Experiments Model Details}

\begin{table}[htbp]
\centering
\small
\caption{The 24 checkpoints included in the main analysis. Gemma3 uses publicly released instruction-tuned checkpoints; therefore, the Gemma2/Gemma3 comparison is not interpreted as a strict base-to-base ablation.}

\begin{tabular}{llc}
\toprule
Model family & Checkpoints & Role in analysis \\
\midrule
Qwen2.5 & 1.5B, 7B, 32B & Dense scaling baseline \\
Qwen2.5-VL & 3B, 7B, 32B & Vision-language adaptation \\
Qwen3 & 1.7B, 8B, 30B-A3B, 32B & Dense/MoE comparison \\
Qwen3.5 & 0.8B, 9B, 27B, 35B-A3B & Low-outlier counterexample family \\
Gemma2 & 2B, 9B, 27B & Non-monotonic scaling case \\
Gemma3 & 4B-it, 27B-it & High-magnitude family \\
Ling & 5T, 10T, 15T, 20T & Training-stage evolution \\
GPT-OSS & 20B & Component-level special case \\
\bottomrule
\end{tabular}

\label{tab:model-families}
\end{table}

\section{Supplementary Main-Text Figures}
\label{app:supplementary-main-figures}

\begin{figure}[htbp]
\centering
\includegraphics[width=.98\linewidth]{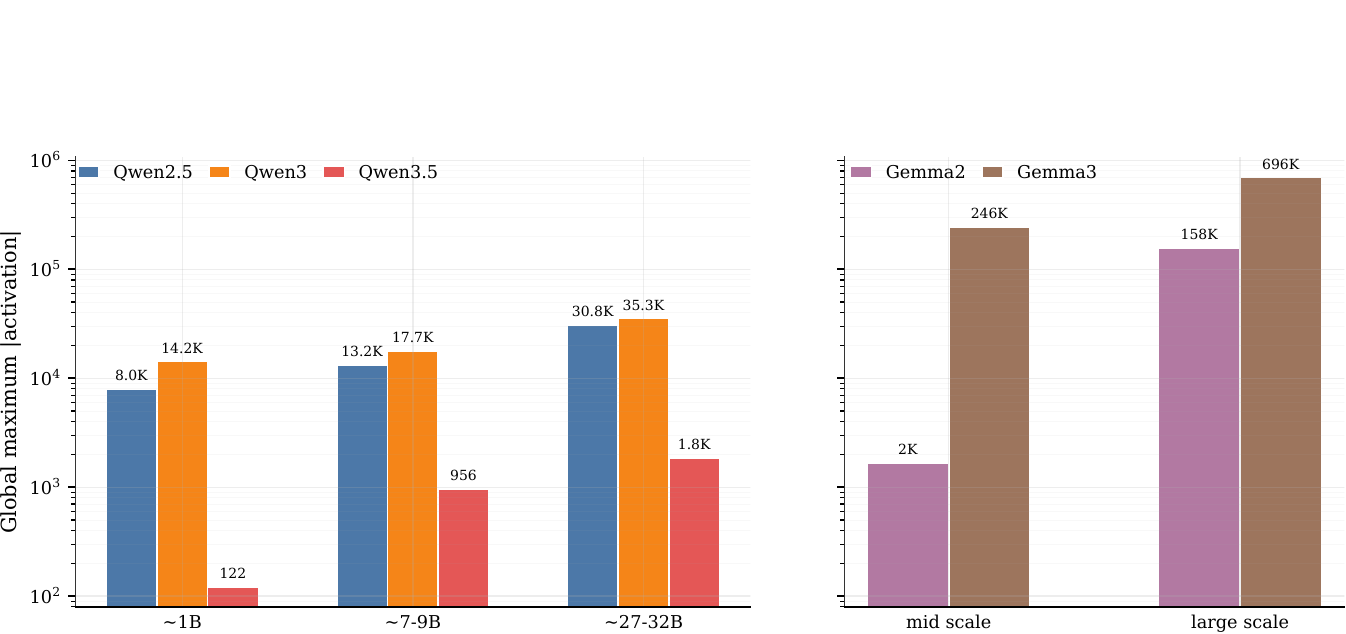}
\caption{Generational evolution at similar sizes. Left: Qwen shows a Qwen2.5$\rightarrow$Qwen3 increase followed by a Qwen3$\rightarrow$Qwen3.5 decrease across three size groups. Right: Gemma2$\rightarrow$Gemma3 increases in both size groups. The vertical axis is global maximum activation magnitude on a log scale.}
\label{fig:generation-same-scale}
\end{figure}

\section{Architectural and Training Factors at Matched Scale}
\label{sec:special-architectures}

This section provides controlled comparisons at identical or similar model scales. We compare MoE versus dense models, vision-language versus text-only checkpoints, base versus instruction-tuned checkpoints, and different Ling-mini training stages.

\subsection{MoE vs. Dense Models}

\begin{figure}[!htbp]
\centering
\includegraphics[width=.76\linewidth]{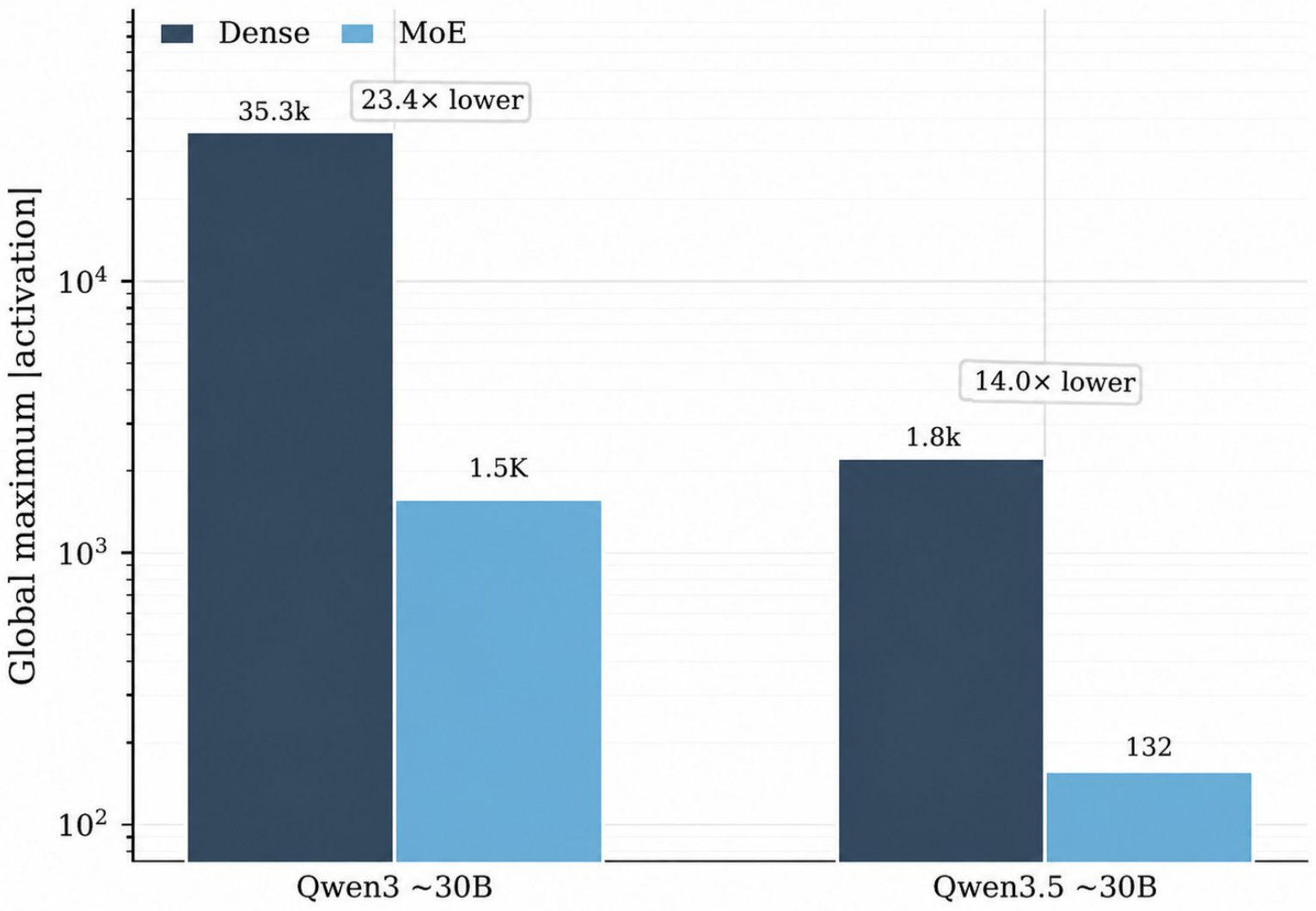}
\caption{Matched-scale comparison of MoE and dense checkpoints. Each bar group fixes model family and approximate total parameter scale while changing only the dense/MoE form. The vertical axis is global maximum activation magnitude on a log scale.}
\label{fig:moe-dense}
\end{figure}

The MoE-versus-dense comparison is the clearest matched-design contrast we have for an architectural axis. Figure~\ref{fig:moe-dense} compares two pairs of same-family checkpoints near the 30B scale. Qwen3-30B-A3B has a global peak of 1,512, which is $23.4\times$ lower than Qwen3-32B at 35,328. Qwen3.5-35B-A3B has a global peak of 132, which is $14.0\times$ lower than the Qwen3.5-27B dense counterpart (the value plotted is the global $M$ across all hooked components and layers, which differs from the representative-layer Top-1 reported in Table~\ref{tab:massive-existence}; see Section~\ref{sec:existence}). The compared checkpoints have similar total parameter counts but MoE models activate fewer parameters per token and route tokens through sparse expert paths. The matched-pair gap is therefore unlikely to be explained by active-parameter count alone; it is consistent with an architectural effect of sparse routing and expert structure. We caution, however, that with only $n=2$ matched pairs and possible unobserved differences in training recipe between paired checkpoints, this contrast is observational rather than causal.

\subsection{Vision-language vs. Text-only models}

\begin{figure}[!htbp]
\centering
\includegraphics[width=.76\linewidth]{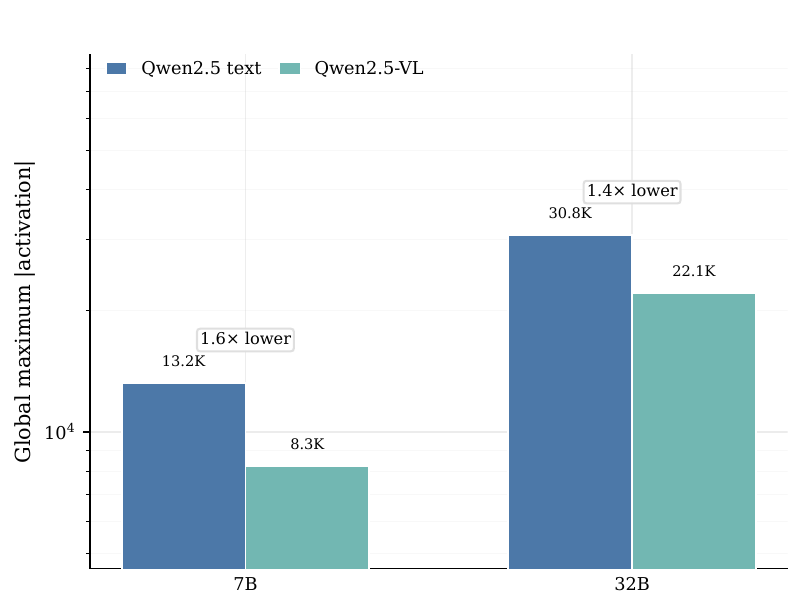}
\caption{Matched-scale comparison between Qwen2.5-VL and text-only Qwen2.5 checkpoints. The two bar groups correspond to 7B and 32B scales. The vertical axis is global maximum activation magnitude on a log scale.}
\label{fig:vl-text}
\end{figure}

Vision-language adaptation does not eliminate extreme activations, but matched-scale checkpoints differ measurably in peak magnitude. Figure~\ref{fig:vl-text} compares Qwen2.5-VL with text-only Qwen2.5 at 7B and 32B. Qwen2.5-VL-7B reaches a global peak of 8,256, $1.6\times$ lower than Qwen2.5-7B at 13,248. At 32B, Qwen2.5-VL reaches 22,144, $1.4\times$ lower than the text-only Qwen2.5-32B counterpart, whose global $M$ is 30,848 across all hooked components and layers (the corresponding Top-1 entry in Table~\ref{tab:massive-existence}, $22{,}144$, is the value at the representative passing-criterion layer; see the Section~\ref{sec:existence} convention note). Vision-language checkpoints therefore remain in a high-magnitude regime comparable to their text-only backbones, but modality adaptation and training-recipe changes co-occur with a moderate same-scale shift in the maximum activation. As with the MoE contrast, this is observational evidence based on $n=2$ matched pairs within a single family.

\subsection{Base vs. Instruct}

\begin{figure}[!htbp]
\centering
\includegraphics[width=.9\linewidth]{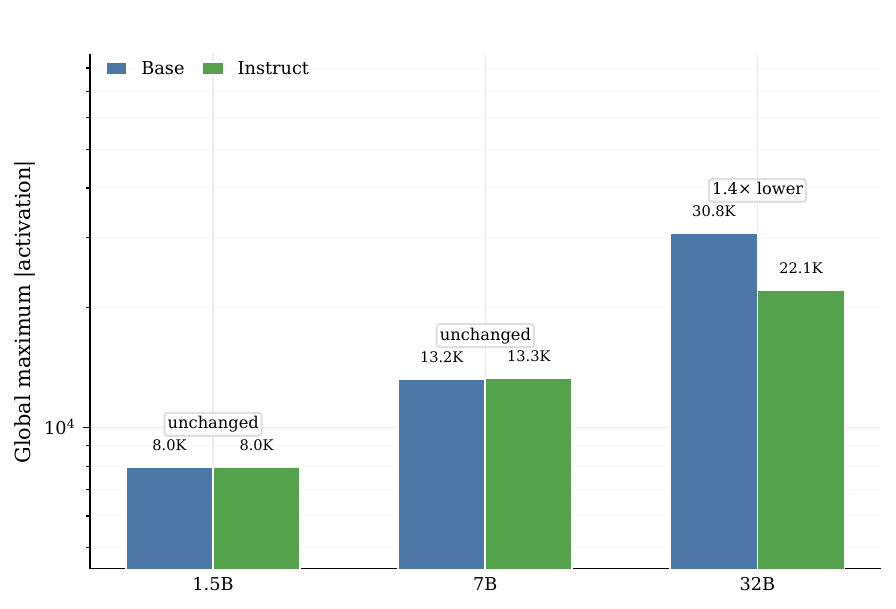}
\caption{Matched-backbone comparison of Qwen2.5 Base and Instruct checkpoints. Each bar group fixes parameter scale and changes only whether the checkpoint has undergone instruction tuning. The vertical axis is global maximum activation magnitude on a log scale.}
\label{fig:base-instruct}
\end{figure}

Instruction tuning affects maximum activations differently from model scaling. Figure~\ref{fig:base-instruct} compares Base and Instruct checkpoints at the same Qwen2.5 backbone and parameter scale. At 1.5B, the global peak remains unchanged at 7,968. At 7B, it increases slightly from 13,248 to 13,312 ($1.005\times$). At 32B, it decreases from 30,848 to 22,144, a $1.4\times$ reduction. The 32B Base value of $30{,}848$ is the global $M$ across all hooked components and layers; the Top-1 entry of $22{,}144$ that appears for Qwen2.5-32B in Table~\ref{tab:massive-existence} is taken at the representative passing-criterion layer, and is by construction $\le M$ (see the Section~\ref{sec:existence} convention note). A closer inspection of layerwise hidden-state peaks shows that SFT mainly changes late layers rather than reorganizing the mid-layer structure. For 1.5B, 7B, and 32B, the stable high-peak middle-layer regions remain at $1.000\times$, $1.005\times$, and $1.000\times$ of their base values, respectively. The final-layer peaks decrease from 1,536 to 840 ($-45\%$), from 4,864 to 2,528 ($-48\%$), and from 30,848 to 21,248 ($-31\%$), with only mild compensation in the penultimate layer. The global-peak decrease in Qwen2.5-32B-Instruct is therefore best understood as late-layer compression that exposes an already existing mid-layer high-peak region as the global maximum, rather than the creation of a new activation structure.

\subsection{Training-stage evolution}

\begin{figure}[!htbp]
\centering
\includegraphics[width=.75\linewidth]{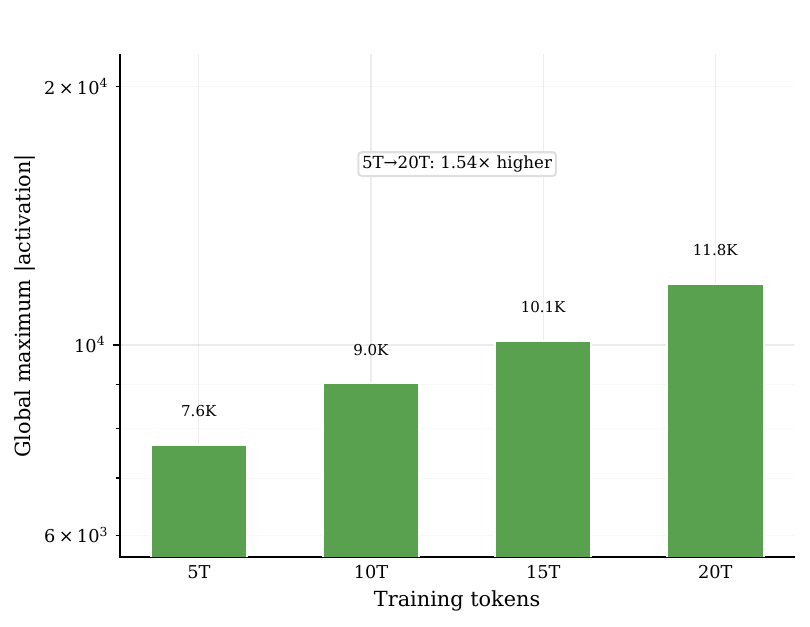}
\caption{Global maximum activation across Ling-mini training stages. The horizontal axis is training tokens, and the vertical axis is global maximum activation magnitude on a log scale.}
\label{fig:ling-training-stage}
\end{figure}

Ling-mini provides a training-stage comparison with fixed family and approximately fixed model scale. Figure~\ref{fig:ling-training-stage} shows that as training increases from 5T to 20T tokens, the global maximum activation rises monotonically from 7,648 to 9,024, then to 9,600, and finally to 10,240, for an overall increase of $1.34\times$. In this training sequence, longer training is associated with larger activation peaks. Even when family, architecture, and parameter scale are approximately held fixed, training progress itself co-occurs with a gradual increase in extreme activation magnitudes. Training stage should therefore be recorded as a separate observation dimension alongside model size and architecture, although the magnitude of this effect ($1.34\times$) is small relative to the cross-family span.

Overall, MoE, vision-language adaptation, SFT, and training stage are all associated with measurable differences in maximum activations, but the contrasts differ in magnitude and in the strength of evidence available. The MoE-vs-dense gap is large (an order of magnitude or more) but rests on $n=2$ matched pairs. The SFT contrast is the strongest matched design ($n=3$ same-backbone sizes) and reveals a layerwise-localized effect (late-layer compression with preserved middle-layer high-peak regions). The vision-language and training-stage contrasts produce moderate within-family shifts. None of these comparisons constitute causal evidence; each could in principle reflect unobserved training-recipe differences between the paired checkpoints.

\section{Discussion}
\label{sec:discussion}

\subsection{Distinction from outlier features}

The object of this study is maximum activation magnitude: the largest absolute activation observed across layers and key components under a unified evaluation corpus. This is a model-level statistic oriented toward dynamic range and deployment risk. It asks how large an activation upper bound a checkpoint may produce during actual forward inference. By contrast, outlier-feature studies often focus on whether certain feature dimensions remain persistently abnormal across many tokens or samples. We therefore do not interpret maximum activation magnitude as a fixed outlier-feature existence test. Instead, we treat it as a direct risk indicator for quantization scales, activation rescaling, and inference stability.

\subsection{Implications for quantization and deployment}

Maximum activation magnitude directly constrains the upper bound of activation scale, and therefore affects scale selection and reconstruction error in low-bit activation quantization. Higher peaks make per-tensor quantization more likely to be dominated by a few extreme values; however, quantization error is also shaped by the full distribution at the peak layer, the effective signal magnitude, and the clipping strategy. Maximum activation magnitude should therefore be treated as an important prior for quantization risk, rather than as the sole determinant of end-to-end quantization quality. Appendix Figure~\ref{fig:tiers} provides a deployment-oriented grouping based on global maximum activation magnitude.

\begin{figure}[!htbp]
\centering
\includegraphics[width=.96\linewidth]{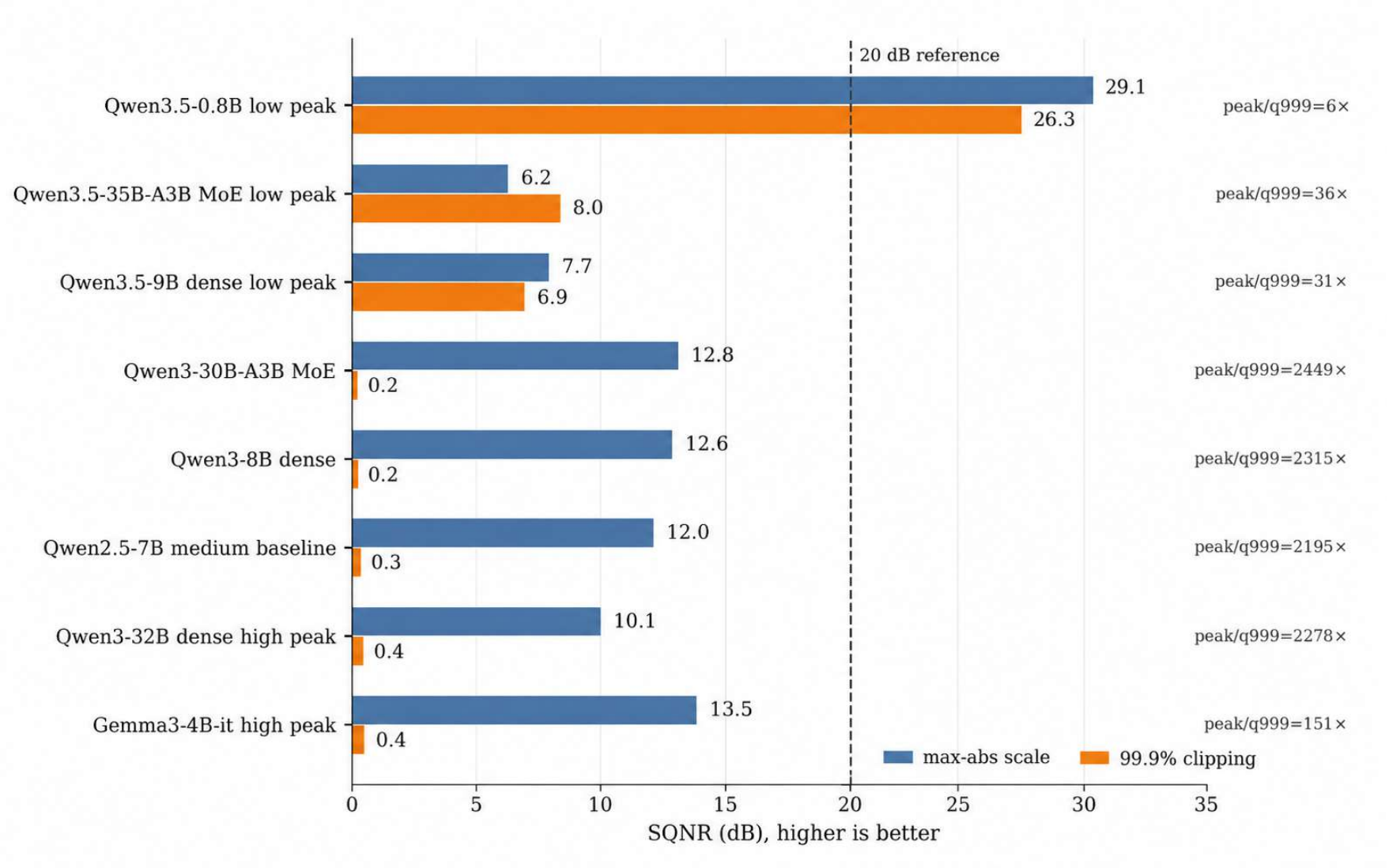}
\caption{INT-8 activation quantization sanity check for eight representative models. Grouped horizontal bars show SQNR at the peak hidden layer under max-abs scaling and 99.9\% clipping. The dashed line marks 20 dB; higher values indicate lower quantization error. Right-side annotations report the ratio between the global peak and the 99.9\% clipping threshold.}
\label{fig:quant-sanity}
\end{figure}

To validate this deployment relevance, we conduct a lightweight INT-8 activation quantization sanity check. The experiment covers eight representative models spanning Qwen2.5, Qwen3, Qwen3.5, Gemma3, and MoE/dense contrasts. For each model, we use 128 samples for calibration and 256 samples for evaluation at the peak hidden layer, and compare two per-tensor symmetric quantization strategies: max-abs scaling and 99.9\% clipping. Figure~\ref{fig:quant-sanity} shows that Qwen3.5-0.8B maintains the highest SQNR under both strategies, at 29.1 dB and 26.3 dB. Most medium- and high-peak models obtain roughly 10--14 dB under max-abs scaling and further drop to approximately 0.2--0.4 dB under 99.9\% clipping. Qwen3.5-9B and Qwen3.5-35B-A3B have relatively low absolute peaks but still show substantially lower SQNR than Qwen3.5-0.8B, indicating that within-layer distributional structure beyond peak magnitude also affects quantization error. Overall, this sanity check demonstrates that maximum activation magnitude can translate into observable activation reconstruction error through scale selection, while also suggesting that end-to-end quantization should be evaluated with richer distributional and task-level measurements.

We deliberately keep this experiment as a \emph{deployment-relevance sanity check} rather than a calibrated dose--response curve. The eight checkpoints are chosen to span the main regimes identified by the survey---low-peak dense (Qwen3.5-0.8B), low-peak Qwen3.5 variants, medium-baseline Qwen2.5, high-peak dense Qwen3 checkpoints, a MoE checkpoint (Qwen3-30B-A3B), and a high-peak Gemma-family checkpoint (Gemma3-4B-it)---so that any $M$-dependence of SQNR has a chance to manifest at the same layer hooks used in the rest of the study. The two recipes (max-abs and $99.9\%$ clipping) bracket common per-tensor symmetric scale-selection strategies and are directly affected by the activation upper bound; advanced mitigations such as rotation-based quantization~\citep{ashkboos2024quarot,liu2024spinquant} or KV-cache prefixing~\citep{chen2024prefixquant} are designed to neutralize $M$ rather than to expose it, so they are intentionally not part of this probe. We therefore read the $M$-versus-SQNR relationship as qualitative evidence that maximum activation magnitude translates into observable scale-selection error, not as a generalized scaling law over all 24 checkpoints.

\textbf{Relation to 2024--2026 quantization advances.} A wave of post-2024 quantization techniques---dual-rotation DuQuant~\citep{lin2024duquant}, learned affine flattening FlatQuant~\citep{sun2025flatquant}, and KV-cache-targeted KIVI~\citep{liu2024kivi}, alongside the earlier rotation-based QuaRot/SpinQuant~\citep{ashkboos2024quarot,liu2024spinquant} and prefix-based PrefixQuant~\citep{chen2024prefixquant}---has substantially narrowed the SQNR gap that motivates max-abs scaling, and a parallel low-precision-pretraining line~\citep{deepseekai2025deepseekv3technicalreport,deepseekv3hardware} folds analogous mitigations into FP8 training itself. These methods \emph{transform away} or \emph{absorb} the activation upper bound; none of them removes it. Our contribution is orthogonal: we measure the upper bound itself across a far wider family/architecture/training-stage span than any of these papers calibrates against, and we show that $M$ varies by  nearly four orders of magnitude across modern open releases. The cost of any of the above mitigations---rotation rank, prefix-token budget, KV-cache bit width, or FP8 block-scale granularity---is itself a monotone function of the dynamic range we report, so the per-checkpoint $M$ values remain a deployment-relevant model card entry even when none of the mitigations are bypassed.

\subsection{Threats to validity and limitations}

The conclusions of this paper should be interpreted within the boundaries of an observational study. We measure maximum activations across open checkpoints under a unified evaluation protocol and analyze their relationships with family, architecture, and training stage. These results reveal stable empirical differences but do not by themselves establish causal training mechanisms. Second, the study covers open LLMs only; closed models or models with unreleased training recipes may exhibit different activation dynamics. Third, maximum activation is an extreme statistic and could be affected by longer contexts, rare inputs, or larger evaluation corpora. Nevertheless, our subsampling experiment shows that representative models reproduce the same peak orders of magnitude across repeated subsamples. Finally, the INT-8 experiment is a deployment-relevance sanity check rather than a complete end-to-end quantization evaluation. Future work could combine intervention experiments, token/feature localization, and end-to-end quantized tasks to explain the mechanisms that create these peaks and determine how controllable they are.

Several further limitations are worth flagging explicitly; these are items that the present manuscript does not address but that we view as the natural next iteration. \textbf{(i) Corpus coverage}. Our 5{,}000-sample evaluation mixture is restricted to English, Chinese, and code domains; activations on long-tail languages, mathematical reasoning chains, or tool-use traces may differ. \textbf{(ii) Context length}. All measurements use sequences up to 4{,}096 tokens, so we cannot speak to whether maxima grow, saturate, or shift carriers under 32k--128k contexts that several of these checkpoints support. \textbf{(iii) Training-stage labels}. Our Base-vs-Instruct contrast treats ``instruction-tuned'' as a single bucket, but the public Instruct checkpoints conflate supervised fine-tuning with downstream RLHF or DPO stages whose training data we do not control; a finer-grained decomposition would require checkpoint releases at each stage. \textbf{(iv) Quantization scope}. The INT-8 sanity check is restricted to $8$ checkpoints, $1$ layer each, and per-tensor recipes only; a 24-point regression of SQNR against $\log_{10} M$, and a comparison against rotation-based~\citep{ashkboos2024quarot,liu2024spinquant} or prefix-based~\citep{chen2024prefixquant} mitigations, would be required to turn the link from $M$ to deployment cost into a calibrated dose--response curve. \textbf{(v) Layerwise-pattern classification}. The two-pattern dichotomy in Section~\ref{sec:layerwise} is identified qualitatively from the depth-normalized heatmap; a quantitative jump-score / clustering diagnostic would be required to convert it into a numerical classifier. \textbf{(vi) Sun-criterion failure mechanisms}. Linking each failing checkpoint (Qwen2.5-1.5B, Qwen3.5-0.8B/-9B/-35B-A3B) to specific normalization or attention dynamics---for example via residual-stream RMS profiles or via the early-layer step-up blocks identified by~\citep{zhu2026spike}---is left to a mechanistic follow-up. \textbf{(vii) Reproducibility artifacts}. The full reproducibility appendix (HuggingFace repository ID and commit revision per checkpoint, dtype, attention implementation, normalization variant, RoPE base, pinned versions of \texttt{torch}/\texttt{transformers}/\texttt{accelerate}, random seeds, code-availability statement) will accompany the public release; the current text describes the pipeline but does not yet pin its concrete artifacts. \textbf{(viii) Sample sizes for matched comparisons}. The MoE-vs-dense and VL-vs-text contrasts each rest on $n=2$ matched pairs within a single family, the training-stage contrast on $1$ family with $4$ stages, and only the Base-vs-Instruct contrast achieves $n=3$; effect-size statements should therefore be read as direction-of-effect rather than as population-level estimates, and additional matched pairs from other families would be required to strengthen them. None of these gaps is fundamental to the measurement framework; they are scoping choices for the present submission, and each is a candidate for the next iteration of this study rather than a planned addition to this manuscript.

\section{Related Work}
\label{sec:related-work}

\paragraph{Outlier features and massive activations.}
Large activations in transformer language models were first brought into focus by low-bit inference failures. \citep{dettmers2022llmint8} showed that sufficiently large OPT- and BLOOM-style models develop sparse, high-magnitude ``emergent features'' whose preservation is necessary for lossless INT8 inference. \citep{sun2024massive} sharpened this observation into the notion of \emph{massive activations}: token-feature coordinates that are both absolutely large and locally sparse relative to the surrounding hidden-state distribution. This line of work established two facts that are now central to activation-aware deployment: the extreme coordinates occupy a tiny fraction of the representation, yet ablating or quantizing them naively can cause disproportionate degradation. Subsequent analyses extended the taxonomy beyond hidden-state activations. \citep{an2025systematicoutliers} relate activation outliers, weight outliers, and attention outliers, arguing that they are mutually structured rather than independent numerical accidents, while \citep{macocco2025heuristic} show that last-layer outlier dimensions can implement useful high-frequency-token prediction heuristics. These studies motivate treating extreme activations as functional model components, not merely as pathological noise.

\paragraph{Attention sinks and mechanistic accounts.}
A major mechanistic explanation links massive activations to the softmax attention constraint. When an attention head has no useful context to retrieve, the probability simplex still forces it to allocate its mass somewhere; models often learn to route this mass to a beginning-of-sequence token or another low-information token, creating an attention sink. \citep{bondarenko2023quantizable} connect such no-op behavior to quantization difficulty, and \citep{gu2025attentionsink} empirically trace when sink behavior emerges during pretraining. More recent work refines this picture. \citep{kaul2025attentionactivation} attribute first-token attention dominance to softmax geometry and large hidden-state kurtosis partly to coordinate-wise adaptive optimization, proposing softmax-1 and OrthoAdam as architectural and optimizer-level remedies. \citep{zhu2026spike} decouple residual-stream activation spikes from local attention sinks, showing that both phenomena co-occur in pre-norm transformers but need not be the same object. \citep{queipo2026attentioncompression} further link attention sinks to compression valleys, proving that extreme residual-stream norms induce representational compression and proposing a mix-compress-refine view of depth-wise computation. Together these works explain why massive activations may be useful for information routing and stabilization, while also clarifying why their magnitude can become a deployment bottleneck.

\paragraph{Quantization and numerical mitigation.}
The practical importance of massive activations is most visible in post-training quantization. Because per-tensor or per-channel scales are often governed by the largest observed magnitude, even a small number of extreme values can waste quantization levels and increase reconstruction error for ordinary activations. LLM.int8() preserves sparse outlier dimensions in higher precision while quantizing the remaining bulk activations \citep{dettmers2022llmint8}. SmoothQuant instead migrates activation difficulty into weights through an algebraically equivalent rescaling \citep{xiao2023smoothquant}, while Outlier Suppression+ shifts and rescales activations to reduce quantization sensitivity \citep{wei2023outliersuppression}. Weight-only and activation-aware methods such as GPTQ and AWQ reduce memory and bandwidth pressure while protecting salient parameters \citep{frantar2023gptq,lin2024awq}. More recent methods transform the representation basis or flatten the activation landscape: QuaRot and SpinQuant use rotations to distribute outlier mass \citep{ashkboos2024quarot,liu2024spinquant}, DuQuant applies dual transformations \citep{lin2024duquant}, FlatQuant learns affine flattening transforms \citep{sun2025flatquant}, and PrefixQuant or KIVI target outliers in prefixed activations and KV caches \citep{chen2024prefixquant,liu2024kivi}. These methods are complementary to our study: they reduce or absorb the numerical effect of large activations, whereas we measure the upper-bound activation magnitude itself across current open model families.

\paragraph{Architectural and multimodal interventions.}
A parallel line of work asks whether outlier-like artifacts arise because transformer architectures lack explicit storage locations for global or null information. In vision transformers, high-norm artifact tokens appear in low-information image patches and harm dense prediction; register tokens provide dedicated learned slots that absorb this global computation and remove the artifact pattern \citep{darcet2024registers}. Post-hoc register methods show that similar benefits can be obtained without full retraining by distilling artifact-free dense features into a lightly modified model \citep{chen2025selfdistilledregisters}. In multimodal language models, \citep{anand2026avsr} show that attention sinks and massive activations can emerge not only at the BOS token but also at intermediate low-semantic audio-visual tokens during fine-tuning, and that decorrelating intermediate tokens from the BOS representation mitigates both the sink and activation effects. These results suggest that maximum activation behavior is sensitive to modality adaptation, token roles, and the availability of architectural ``scratch space.''

\paragraph{Position of this work.}
Existing studies have primarily treated massive activations as a binary phenomenon, a mechanistic artifact, or an obstacle to be removed by a particular quantization method. They also concentrate on a limited set of earlier LLaMA-, OPT-, or BLOOM-style checkpoints. Modern open models now vary along many additional axes, including normalization and training recipes, dense versus MoE computation, multimodal adaptation, instruction tuning, and released intermediate training stages \citep{qwen25report,qwen3report,qwen25vl,gemma2report,gemma3report,lingreport,gptoss2025}. Our work is therefore complementary to both the interpretability and quantization literatures. Rather than asking only whether a model contains a massive activation under a fixed local criterion, we measure the continuous deployment-relevant statistic $M=\max |a|$ under a unified protocol, compare it across recent open families and matched architectural or training contrasts, and connect it to activation-quantization reconstruction error. This framing turns maximum activation magnitude into a model-level property that should be reported alongside open-weight releases.

\section{Appendix Summary}
\label{sec:conclusion}

We presented a unified measurement study of maximum activations across 24 checkpoints from 8 modern open LLM families. Our central message is empirical and definitional: \emph{the maximum activation magnitude $M$ of an open LLM is a continuous, releasable model property that is not predicted by parameter count alone}. Within a family, $M$ often grows with model size, but cross-family, cross-generation, and cross-architecture comparisons break simple monotonic scaling, and matched-design contrasts associate MoE routing, vision-language adaptation, supervised fine-tuning, and training stage with measurable shifts in $M$ or in its layerwise structure. The residual stream carries the global maximum in 22 of 24 checkpoints, with GPT-OSS-20B as a single MLP-output exception that any quantization scheme should accommodate. A lightweight INT-8 sanity check further indicates that $M$ co-varies with low-bit reconstruction error through its effect on activation-scale selection. We therefore recommend that $M$ and its layerwise carrier be reported alongside any open-weight release as part of a deployment-oriented model card extension. We deliberately stop short of causal claims about why families differ; explaining the mechanism behind these differences---and intervening on it---is left as future work.

\section{Additional Figures}
\label{app:additional-figures}

The appendix keeps only figures that complement the main-text conclusions and provide additional verification value. We omit separate plots for local-sparsity pass rates and per-family global peaks because their information overlaps with the main text and with Figures~\ref{fig:peak-landscape} and~\ref{fig:family-scaling-non-moe}. The retained figures focus on three types of evidence: deployment-oriented magnitude tiers, per-family layerwise trajectories, and component-level carrier differences in representative models.

\subsection{Deployment-oriented tiers}
\label{app:deployment-tiers}

\begin{figure}[!htbp]
\centering
\includegraphics[width=.92\linewidth]{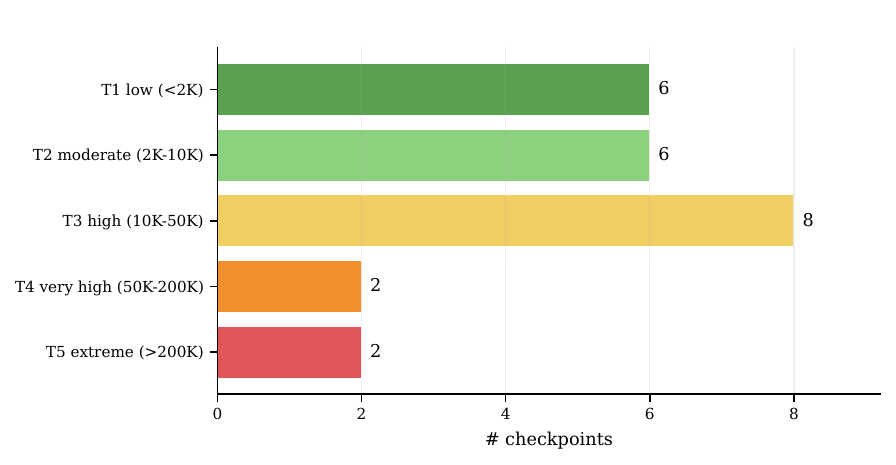}
\caption{Deployment-oriented tiers based on global maximum activation magnitude. The horizontal bar chart counts checkpoints in each tier. The horizontal layout avoids crowded tier labels. The figure is based only on observed activation magnitudes and is intended to indicate that different models may require different quantization treatment.}
\label{fig:tiers}
\end{figure}

Figure~\ref{fig:tiers} groups all main-analysis checkpoints into five order-of-magnitude tiers according to global maximum activation magnitude. The figure is not intended to propose a new threshold standard; rather, it provides an intuitive deployment-oriented reference for selecting quantization and activation-scaling strategies.

\subsection{Per-family layerwise maximum activations}
\label{app:per-family-layerwise-max}

\begin{figure}[!htbp]
\centering
\includegraphics[width=.98\linewidth]{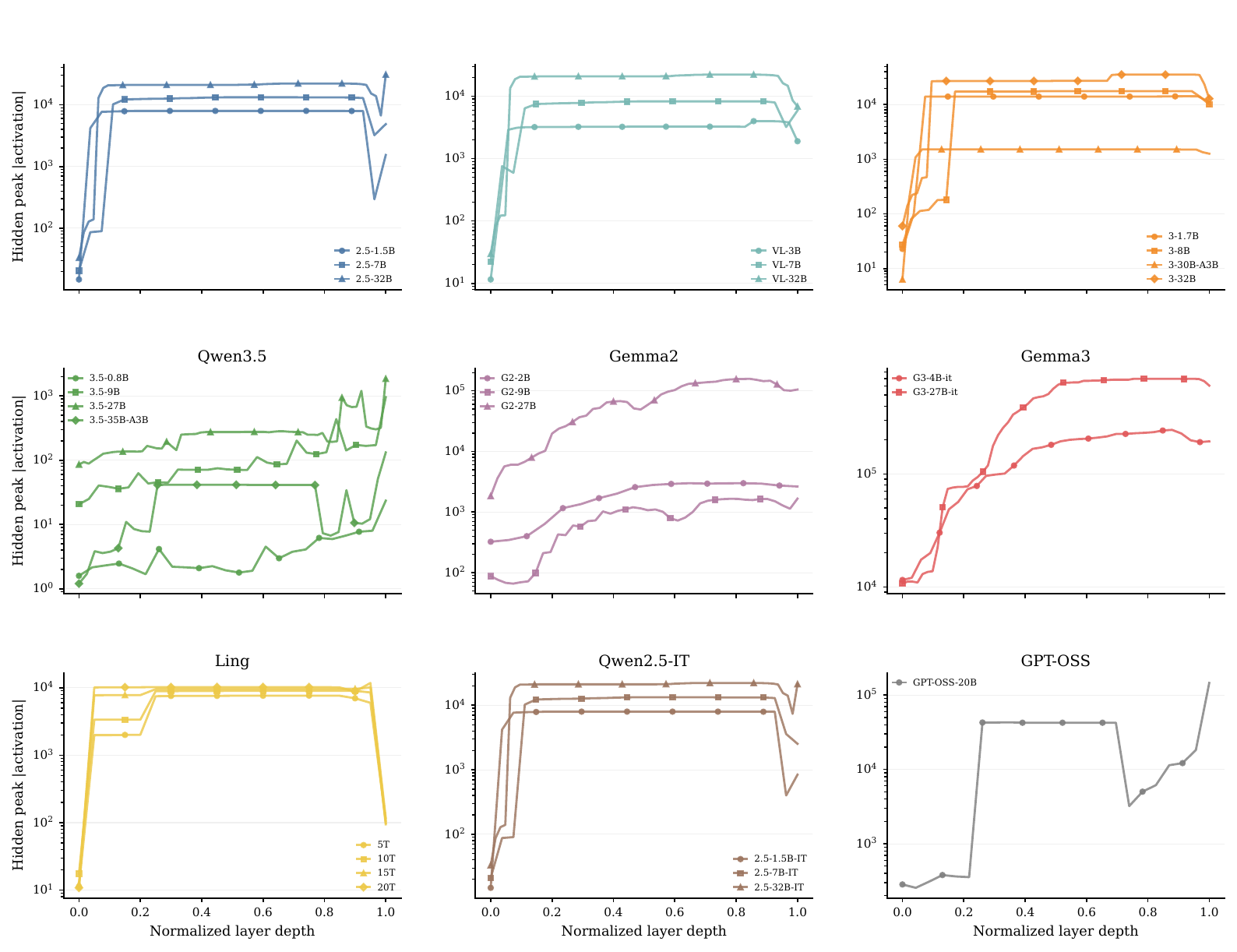}
\caption{Hidden-state layerwise maximum-activation trajectories within each model family. Each subplot corresponds to a family or training form, and each curve corresponds to one checkpoint in that group. The vertical axis uses a log scale to accommodate cross-family magnitude differences.}
\label{fig:app-layerwise-family-grid}
\end{figure}

Figure~\ref{fig:app-layerwise-family-grid} shows hidden-state layerwise maximum-activation trajectories within each model family. The horizontal axis is normalized depth, and the vertical axis is hidden-state layerwise maximum absolute activation. Because peak magnitudes span several orders of magnitude across families, the vertical axis uses a log scale.

\subsection{Component-level maximum activation distributions}
\label{app:component-layerwise-max}

\begin{figure}[!htbp]
\centering
\includegraphics[width=.95\linewidth]{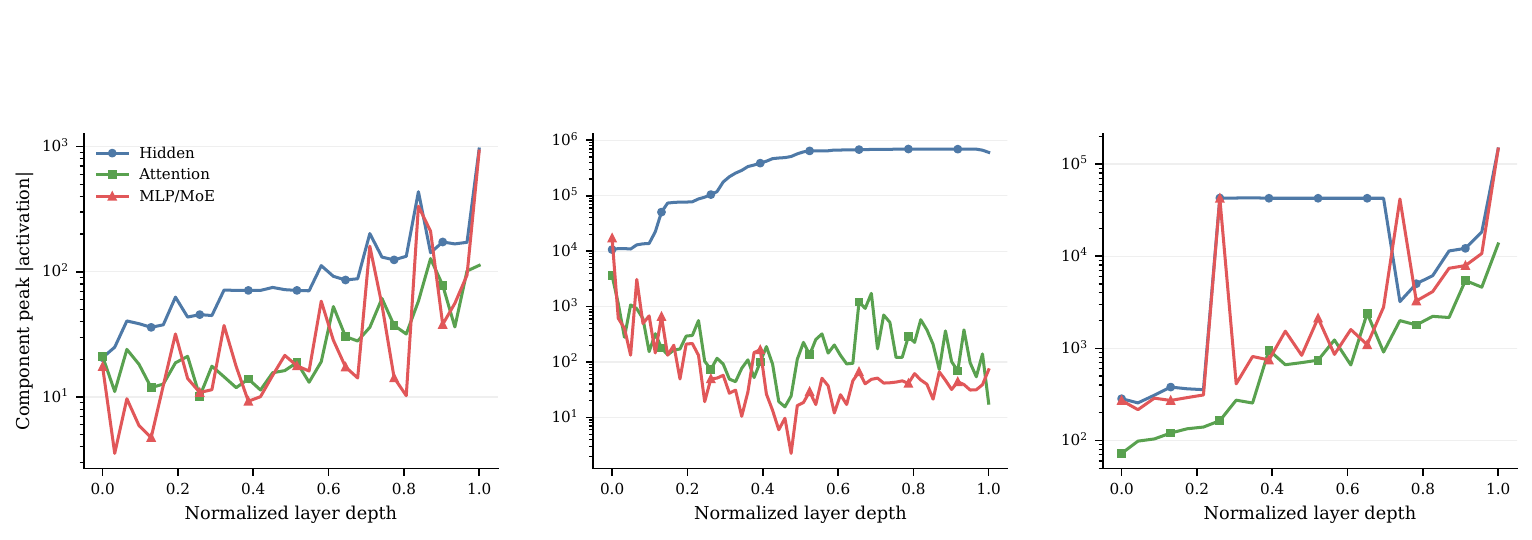}
\caption{Component-level maximum-activation trajectories for representative models. The three subplots show a low-peak model, a high-peak model, and the GPT-OSS component-level exception. Curves correspond to hidden states, attention outputs, and MLP/MoE outputs; the vertical axis is component-level maximum absolute activation on a log scale.}
\label{fig:app-component-representatives}
\end{figure}

Figure~\ref{fig:app-component-representatives} presents component-level layerwise trajectories for three representative checkpoints, complementing the main-text discussion of carrier components. The three subplots correspond to a low-peak model, a high-peak model, and the GPT-OSS component-level exception. Compared with plotting all checkpoints, these representative panels more clearly show the magnitude differences among hidden states, attention outputs, and MLP/MoE outputs.

\end{document}